%% file: MICCAI2023.tex
\documentclass[runningheads]{llncs}
\usepackage{graphicx}
\usepackage{amsmath}
\usepackage{multicol}
\usepackage{multirow}
\usepackage{amssymb}
\usepackage{amsfonts}
\usepackage{booktabs}
\usepackage{siunitx}
\usepackage{hyperref}
\usepackage{marvosym}
\begin{document}
\title{Toward Fairness Through Fair Multi-Exit Framework for Dermatological Disease Diagnosis}
\titlerunning{Toward Fairness Through Fair Multi-Exit Framework}
%

\author{Ching-Hao Chiu\inst{1*} \and
Hao-Wei Chung\inst{1*} \and
Yu-Jen Chen\inst{1} \and Yiyu Shi\inst{2} \and Tsung-Yi Ho\inst{3} }
\authorrunning{C. Chiu, H. Chung et al.}
%
\institute{National Tsing Hua University, Taiwan \\
 \email{ \{gwjh101708, xdmanwww, yujenchen\}@gapp.nthu.edu.tw} \and
University of Notre Dame, Notre Dame, IN, USA \\
\email{yshi4@nd.edu}\\ \and
The Chinese University of Hong Kong, Hong Kong \\
\email{tyho@cse.cuhk.edu.hk} \\
* Equal contributions. Listing order determined by coin flipping.}

\maketitle              
\begin{abstract}
\label{sec:abstract}
\input{Section/Abstract.tex}
\end{abstract}

\section{Introduction}
\label{sec:intro}
\input{Section/Introduction}

\section{Motivation}
\label{sec:motivation}
\input{Section/Motivation}

\section{Method}
\input{Section/Method}

\section{Experiment}
\input{Section/Experiments}

\section{Results}
\label{sec:result}
\input{Section/Results}

\section{Conclusion}
\input{Section/Conclusion}

{\small
\bibliographystyle{splncs04}
\bibliography{MICCAI,vincent_paper}
}

%
%
%
%




\end{document}


\begin{figure}
\begin{center}

\includegraphics[width=0.95\linewidth]{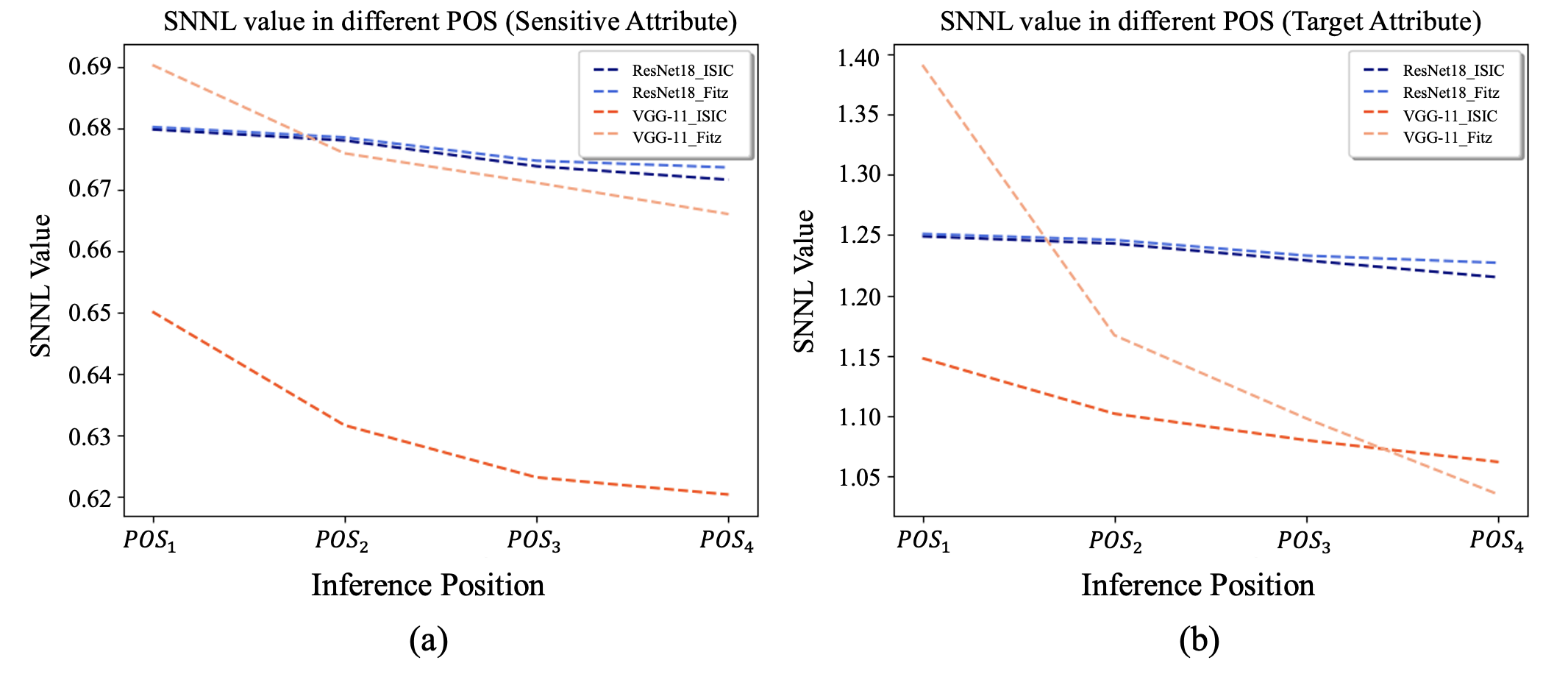}
\caption{
The figure depicts the experiment in the Motivation Section, which shows the SNNL of sensitive and target attributes at different inference positions for ResNet18 and VGG-11. ResNet18 and VGG-11 extract internal features from the end of each residual block and the last four max pooling layers, respectively. The $POS$ refers to the inference position in the network from shallow to deep. The results indicate a consistent decrease in the SNNL for sensitive and target attributes from shallow to deep layers across both datasets, which suggests that features are becoming more distinguishable concerning both sensitive and target attributes (accurate but unfair).
}
\label{Fig.Motivation_supp}
\end{center}
\end{figure}

\input{Table/supp_compare_fitz_me}

%% file: Section/Abstract.tex
Fairness has become increasingly pivotal in medical image recognition. However, without mitigating bias, deploying unfair medical AI systems could harm the interests of underprivileged populations. In this paper, we observe that while features extracted from the deeper layers of neural networks generally offer higher accuracy, fairness conditions deteriorate as we extract features from deeper layers. This phenomenon motivates us to extend the concept of multi-exit frameworks. Unlike existing works mainly focusing on accuracy, our multi-exit framework is fairness-oriented; the internal classifiers are trained to be more accurate and fairer, with high extensibility to apply to most existing fairness-aware frameworks. During inference, any instance with high confidence from an internal classifier is allowed to exit early. Experimental results show that the proposed framework can improve the fairness condition over the state-of-the-art in two dermatological disease datasets.



%% file: Section/Introduction.tex
In recent years, machine learning-based \cite{chen2020zero,chen2021one,chen2022representative,chen2021ct,wen2020noises} medical diagnosis systems have been introduced by many institutions. Although these systems achieve high accuracy in predicting medical conditions, bias has been found in dermatological disease datasets as shown in \cite{groh2021evaluating,kinyanjui2020fairness,tschandl2018ham10000}. This bias can arise when there is an imbalance in the number of images representing different skin tones, which can lead to inaccurate predictions and misdiagnosis due to biases towards certain skin tones. The discriminatory nature of these models can have a negative impact on society by limiting access to healthcare resources for different sensitive groups, such as those based on race or gender.


Several methods are proposed to alleviate the bias in machine learning models, including pre-processing, in-processing, and post-processing strategies. Pre-processing strategies adjust training data before training \cite{lu2020gender,ngxande2020bias} or assign different weights to different data samples to suppress the sensitive information during training \cite{kamiran2012data}. In-processing modifies the model architecture, training strategy, and loss function to achieve fairness, such as adversarial training \cite{alvi2018turning,zhang2018mitigating} or regularization-based \cite{jung2021fair,quadrianto2019discovering} methods. Recently, pruning \cite{wu2022fairprune} techniques have also been used to achieve fairness in dermatological disease diagnosis. However, these methods may decrease accuracy for both groups and do not guarantee explicit protection for the unprivileged group when enforcing fairness constraints. Post-processing techniques enhance fairness by adjusting the model's output distribution. This calibration is done by taking the model's prediction, and the sensitive attribute as inputs \cite{du2020fairness,hardt2016equality,zhao2017men}. However, pre-processing and post-processing methods have limitations that are not applicable to dermatological disease diagnostic tasks since they need extra sensitive information during the training time.

In this paper, we observe that although features from a deep layer of a neural network are discriminative for target groups (i.e., different dermatological diseases), they cause fairness conditions to deteriorate, and we will demonstrate this observation by analyzing the entanglement degree regarding sensitive information with the soft nearest neighbor loss \cite{frosst2019analyzing} of image features in Section \ref{sec:motivation}. This finding is similar to ``overthinking'' phenomenon in neural networks \cite{kaya2019shallow}, where accuracy decreases as the features come from deeper in a neural network and motivate us to use a multi-exit network \cite{kaya2019shallow,teerapittayanon2016branchynet} to address the fairness issue.


Through extensive experiments, we demonstrate that our proposed multi-exit convolutional neural network (ME-CNN) can achieve fairness without using sensitive attributes (unawareness) in the training process, which is suitable for dermatological disease diagnosis because the sensitive attributes information exists privacy issues and is not always available. We compare our approach to the current state-of-the-art method proposed in \cite{wu2022fairprune} and find that the ME-CNN can achieve similar levels of fairness. To further improve fairness, we designed a new framework for fair multi-exit. With the fairness constraint and the early exit strategy at the inference stage, a sufficient discriminative basis can be obtained based on low-level features when classifying easier samples. This contributes to selecting a more optimal prediction regarding the trade-off between accuracy and fairness for each test instance.



The main contributions of the proposed method are as follows: 

\begin{itemize}
    \item Our quantitative analysis shows that the features from a deep layer of a neural network are highly discriminative yet cause fairness to deteriorate.
    \item We propose a fairness through unawareness framework and use multi-exit training to improve fairness in dermatological disease classification.
    \item We demonstrate the extensibility of our framework, which can be applied to various state-of-the-art models and achieve further improvement. Through extensive experiments, we show that our approach can improve fairness while keeping competitive accuracy on both the dermatological disease dataset, ISIC 2019 \cite{combalia2019bcn20000}, and Fitzpatrick-17k datasets \cite{groh2021evaluating}. 

\end{itemize}

%% file: Section/Motivation.tex
In this section, we will discuss the motivation behind our work. The soft nearest neighbor loss (SNNL), as introduced in \cite{frosst2019analyzing}, measures the entanglement degree between features for different classes in the embedded space, and can serve as a proxy for analyzing the degree of fairness in a model. Precisely, we measure the features of SNNL concerning the different sensitive attributes. When the measured SNNL is high, the entangled features are indistinguishable among sensitive attributes. On the other hand, when the SNNL is low, the feature becomes more distinguishable between the sensitive attributes, leading to a biased performance in downstream tasks since the features consist of sensitive information.

To evaluate the SNNL, we analyzed the performance of ResNet18 \cite{he2016deep} and VGG-11 \cite{simonyan2014very} on the ISIC 2019 and Fitzpatrick-17k datasets. We compute the SNNL of sensitive attributes at different inference positions and observed that the SNNL in both datasets decreased by an average of 1.1\% and 0.5\%, respectively, for each inference position from shallow to deep in ResNet18. For VGG-11, the SNNL in both datasets decreases by an average of 1.4\% and 1.2\%, respectively. This phenomenon indicates that the features become more distinguishable to sensitive attributes. The details are provided in the supplementary materials.


A practical approach to avoiding using features that are distinguishable to sensitive attributes for prediction is to choose the result at a shallow layer for the final prediction. To the best of our knowledge, we are the first work that leverages the multi-exit network to improve fairness. In Section \ref{sec:result}, we demonstrate that our framework can be applied to different network architectures and datasets.

%% file: Section/Method.tex
\begin{figure*}[ht]
\begin{center}

\includegraphics[width=0.9\linewidth]{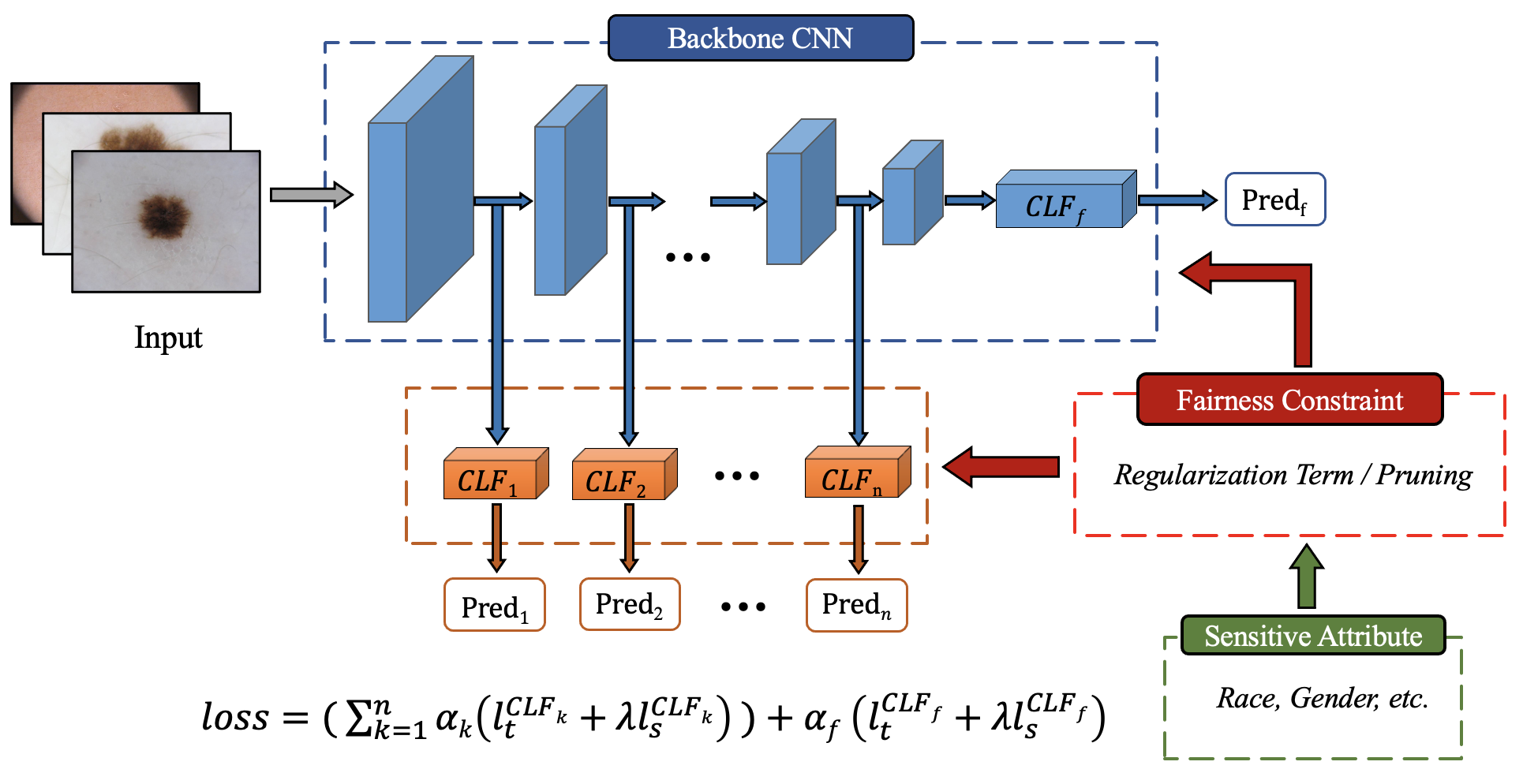}
\caption{Illustration of the proposed multi-exit training framework. $l_t$ and $l_s$ are the loss function related to target and sensitive attributes, respectively. The classifiers $CLF_{1}$ through $CLF_{n}$ refer to internal classifiers, while $CLF_{f}$ refers to the original final classifier.
}
\label{Fig.Method}
\end{center}
\end{figure*}

\subsection{Problem Formulation}
\label{sec:method_problem}

In the classification task, define input features $ x \in X = \mathbb{R}^{d}$,
target class, $ y \in Y = \{1,2, ..., N \} $, and sensitive attributes $a \in A = \{1,2, ..., M \} $. In this paper, we focus on the sensitive attributes in binary case, that is $a \in A = \{0, 1\} $. The goal is to learn a classifier $f : X \rightarrow Y$ that predicts the target class $y$ to achieve high accuracy while being unbiased to the sensitive attributes $a$. 


\subsection{Multi-Exit (ME) Training Framework}
\label{sec:method_main}

Our approach is based on the observation that deep neural networks can exhibit bias against certain sensitive groups, despite achieving high accuracy in deeper layers. To address this issue, we propose a framework leveraging an early exit policy, which allows us to select a result at a shallower layer with high confidence while maintaining accuracy and mitigating fairness problems.


We illustrate our multi-exit framework in Figure \ref{Fig.Method}. Our proposed loss function consists of the cross-entropy loss, $l_t$, and a fairness regularization loss, $l_s$, such as the Maximum Mean Discrepancy (MMD) \cite{jung2021fair} or the Hilbert-Schmidt Independence Criterion (HSIC) \cite{quadrianto2019discovering}, which are replicated for each internal classifier ($CLF_n$) and original final classifier ($CLF_f$). The final loss is obtained through a weighted sum of each $CLF$'s loss, i.e., $loss = (\sum_{k=1}^{n} \alpha _{k}(l_{t}^{CLF_{k}} + \lambda l_{s}^{CLF_{k}})) + \alpha _{f}(l_{t}^{CLF_{f}} + \lambda l_{s}^{CLF_{f}})$, where $\alpha$ is determined by the depth of each $CLF$, similar to \cite{kaya2019shallow,lee2015deeply}. Moreover, the hyperparameter $\lambda$ controls the trade-off between accuracy and fairness. Our approach ensures that the model optimizes for accuracy and fairness, leveraging both shallow and deep layer features in the loss function. Even without any fairness regularization ($\lambda=0$), our experiments demonstrate a notable improvement in fairness (see Section \ref{ME_different_baseline}). 

Our framework can also be extended to other pruning-based fairness methods, such as FairPrune \cite{wu2022fairprune}. We first optimize the multi-exit model using the original multi-exit loss function and then prune it using the corresponding pruning strategy. Our approach has been shown to be effective (see Section \ref{Compare_SOTA}).

During inference, we calculate the softmax score of each internal classifier’s prediction, taking the maximum probability value as the confidence score. We use a confidence threshold $\theta$ to maintain fairness and accuracy. High-confidence instances exit early, and we select the earliest internal classifier with confidence above $\theta$ for an optimal prediction of accuracy and fairness.



%% file: Section/Experiments.tex
\subsection{Dataset} 
In this work, we evaluate our method on two dermatological disease datasets, including ISIC 2019 challenge \cite{combalia2019bcn20000,tschandl2018ham10000} and the Fitzpatrick-17k dataset \cite{groh2021evaluating} . ISIC 2019 challenge contains 25,331 images in 9 diagnostic categories for target labels, and we take gender as our sensitive attribute. The Fitzpatrick-17k dataset contains 16,577 images in 114 skin conditions of target labels and defines skin tone as the sensitive attribute. Next, we apply the data augmentation, including random flipping, rotation, scaling and autoaugment \cite{cubuk2018autoaugment}. After that, we follow the same data split described in \cite{wu2022fairprune} to split the data.


\subsection{Implementation Details and Evaluation Protocol}


We employ ResNet18 and VGG-11 as the backbone architectures for our models. The baseline CNN and the multi-exit models are trained for 200 epochs using an SGD optimizer with a batch size of 256 and a learning rate of 1e-2. Each backbone consisted of four internal classifiers ($CLF$s) and a final classifier ($CLF_{f}$). For ResNet18, the internal features are extracted from the end of each residual block, and for VGG-11 the features are extracted from the last four max pooling layers. The loss weight hyperparameter $\alpha$ is selected based on the approach of \cite{kaya2019shallow} and set to $[0.3, 0.45, 0.6, 0.75, 0.9]$ for the multi-exit models and the $\lambda$ is set to 0.01. The confidence threshold $\theta$ of the test set is set to 0.999, based on the best result after performing a grid search on the validation set.

To evaluate the fairness performance of our framework, we adopted the multi-class equalized opportunity (Eopp0 and Eopp1) and equalized odds (Eodd) metrics proposed in \cite{hardt2016equality}. Specifically, we followed the approach of \cite{wu2022fairprune} for calculating these metrics.

\input{Table/compare_isic}

%% file: Table/compare_isic.tex
\begin{table}[ht]
\centering
\caption{Results of accuracy and fairness of different methods on ISIC 2019 dataset, using gender as the sensitive attribute. The female is the privileged group with higher accuracy by vanilla training. All results are our implementation. }
\setlength{\tabcolsep}{6pt}
\label{Table:compare_isic}
\scriptsize
\begin{tabular}{c c c c c c c c}\toprule 
\multicolumn{1}{c }{\text{}} & \multicolumn{1}{c }{\text{}} &\multicolumn{3}{c }{\text{Accuracy}} & \multicolumn{3}{c }{\text{Fairness}}  \\ 
\cmidrule(lr){3-5}
\cmidrule(ll){6-8}
\multicolumn{1}{ c }{\text{Method}} & \text{Gender} &\text{Precision} & \text{Recall} & \text{F1-score} & \text{Eopp0 $\downarrow$} & \text{Eopp1 $\downarrow$} & \text{Eodd $\downarrow$} \\ 
\hline
\hline
 {\multirow{4}{*}{ResNet18}} & \text{Female} & 0.793 & 0.721 & 0.746 & \multirow{4}{*}{0.006} & \multirow{4}{*}{0.044} & \multirow{4}{*}{0.022} \\ 
 & \text{Male} & 0.731 & 0.725 & 0.723 &  &  & \\ 
 & \text{Avg. $\uparrow$} & 0.762 & 0.723 & 0.735 &  &  & \\ 
 & \text{Diff. $\downarrow$} & 0.063 & 0.004 & 0.023 &  &  & \\ 

\hline

\multirow{4}{*}{AdvConf} & \text{Female} & 0.755 & 0.738 & 0.741 & \multirow{4}{*}{0.008} & \multirow{4}{*}{0.070} & \multirow{4}{*}{0.037} \\ 
 & \text{Male} & 0.710 & 0.757 & 0.731 &  &  & \\ 
 & \text{Avg. $\uparrow$} & 0.733 & 0.747 & 0.736 &  &  & \\ 
 & \text{Diff. $\downarrow$} & 0.045 & 0.020 & 0.010 &  &  & \\ 
\hline
\multirow{4}{*}{AdvRev} & \text{Female} & 0.778 & 0.683 & 0.716 & \multirow{4}{*}{0.007} & \multirow{4}{*}{0.059} & \multirow{4}{*}{0.033} \\ 
 & \text{Male} & 0.773 & 0.706 & 0.729 &  &  & \\ 
 & \text{Avg. $\uparrow$} & 0.775 & 0.694 & 0.723 &  &  & \\ 
 & \text{Diff. $\downarrow$} & 0.006 & 0.023 & 0.014 &  &  & \\ 
\hline
\multirow{4}{*}{DomainIndep} & \text{Female} & 0.729 & 0.747 & 0.734 & \multirow{4}{*}{0.010} & \multirow{4}{*}{0.086} & \multirow{4}{*}{0.042} \\ 
 & \text{Male} & 0.725 & 0.694 & 0.702 &  &  & \\ 
 & \text{Avg. $\uparrow$} & 0.727 & 0.721 & 0.718 &  &  & \\ 
 & \text{Diff. $\downarrow$} & 0.004 & 0.053 & 0.031 &  &  & \\ 
\hline
\multirow{4}{*}{HSIC} & \text{Female} & 0.744 & 0.660 & 0.696 & \multirow{4}{*}{0.008} & \multirow{4}{*}{0.042} & \multirow{4}{*}{0.020} \\ 
 & \text{Male} & 0.718 & 0.697 & 0.705 &  &  & \\ 
 & \text{Avg. $\uparrow$} & 0.731 & 0.679 & 0.700 &  &  & \\ 
 & \text{Diff. $\downarrow$} & 0.026 & 0.037 & 0.009 &  &  & \\ 

\hline
\multicolumn{1}{ c }{\multirow{4}{*}{MFD}} & \text{Female} & 0.770 & 0.697 & 0.726 & \multirow{4}{*}{\textbf{0.005}} & \multirow{4}{*}{0.051} & \multirow{4}{*}{0.024} \\ 
 & \text{Male} & 0.772 & 0.726 & 0.744 &  &  & \\ 
 & \text{Avg. $\uparrow$} & 0.771 & 0.712 & 0.735 &  &  & \\ 
 & \text{Diff. $\downarrow$} & 0.002 & 0.029 & 0.018 &  &  & \\ 

\hline
\hline
{\multirow{4}{*}{FairPrune}} & \text{Female} & 0.776 & 0.711 & 0.734 & \multirow{4}{*}{0.007} & \multirow{4}{*}{0.026} & \multirow{4}{*}{0.014} \\ 
 & \text{Male} & 0.721 & 0.725 & 0.720 &  &  & \\ 
 & \text{Avg. $\uparrow$} & 0.748 & 0.718 & 0.727 &  &  & \\ 
 & \text{Diff. $\downarrow$} & 0.055 & 0.014 & 0.014 &  &  & \\ 
\hline
\multicolumn{1}{ c }{\multirow{4}{*}{ME-FairPrune}} & \text{Female} & 0.770 & 0.723 & 0.742 & \multirow{4}{*}{0.006} & \multirow{4}{*}{\textbf{0.020}} & \multirow{4}{*}{\textbf{0.010}} \\ 
 & \text{Male} & 0.739 & 0.728 & 0.730 &  &  & \\ 
 & \text{Avg. $\uparrow$} & 0.755 & 0.725 & 0.736 &  &  & \\ 
 & \text{Diff. $\downarrow$} & 0.032 & 0.005 & 0.012 &  &  & \\ 
\bottomrule
\end{tabular}
\end{table}

%% file: Section/Results.tex
\subsection{Comparison with State-of-the-art}
\label{Compare_SOTA}
\input{Table/compare_fitz}

In this section, we compare our framework with several baselines, including CNN (ResNet18 and VGG-11), AdvConf \cite{alvi2018turning}, AdvRev \cite{zhang2018mitigating}, DomainIndep \cite{wang2020towards}, HSIC \cite{quadrianto2019discovering}, and MFD \cite{jung2021fair}. We also compare our framework to the current state-of-the-art method FairPrune \cite{wu2022fairprune}. For each dataset, we report accuracy and fairness results, including precision, recall, and F1-score metrics.

\textbf{ISIC 2019 Dataset.}
In Table \ref{Table:compare_isic}, ME-FairPrune refers to the FairPrune applied in our framework. It shows that our ME framework has improved all fairness scores and accuracy in all average scores when applied to FairPrune. Additionally, the difference in each accuracy metric is smaller than that of the original FairPrune. This is because the ME framework ensures that the early exited instances have a high level of confidence in their correctness, and the classification through shallower, fairer features further improves fairness. 

\textbf{Fitzpatrick-17k Dataset.}
To evaluate the extensibility of our framework on different model structures, we use VGG-11 as the backbone of the Fitzpatrick-17k dataset. Table \ref{Table:compare_fitz} demonstrates that the ME framework outperforms all other methods with the best Eopp1 and Eodd scores, showing a 7.5\% and 7.9\% improvement over FairPrune, respectively. Similar to the ISIC 2019 dataset, our results show better mean accuracy and more minor accuracy differences in all criteria than FairPrune. Furthermore, the F1-score and Precision average value are superior to other methods, which show 8.1\% and 4.1\% improvement over FairPrune, respectively.

\input{Table/compare_isic_me}
\subsection{Multi-Exit Training on Different Method}
\label{ME_different_baseline}
In this section, we evaluate the performance of our ME framework when applied to different methods. Table \ref{Table:compare_isic_me} presents the results of our experiments on ResNet18, MFD, and HSIC. Our ME framework improved the Eopp1 and Eodd scores of the original ResNet18 model, which did not apply fairness regularization loss, $l_s$, in total loss. Furthermore, our framework achieved comparable performance in terms of fairness to FairPrune, as shown in Table \ref{Table:compare_isic}. This demonstrates its potential to achieve fairness without using sensitive attributes during training.

We also applied our ME framework to MFD and HSIC, which initially exhibited better fairness performance than other baselines. With our framework, these models showed better fairness while maintaining similar levels of accuracy. These findings suggest that our ME framework can improve the fairness of existing models, making them more equitable without compromising accuracy.


\subsection{Ablation Study}




\begin{figure}
\begin{center}

\includegraphics[width=0.93\linewidth]{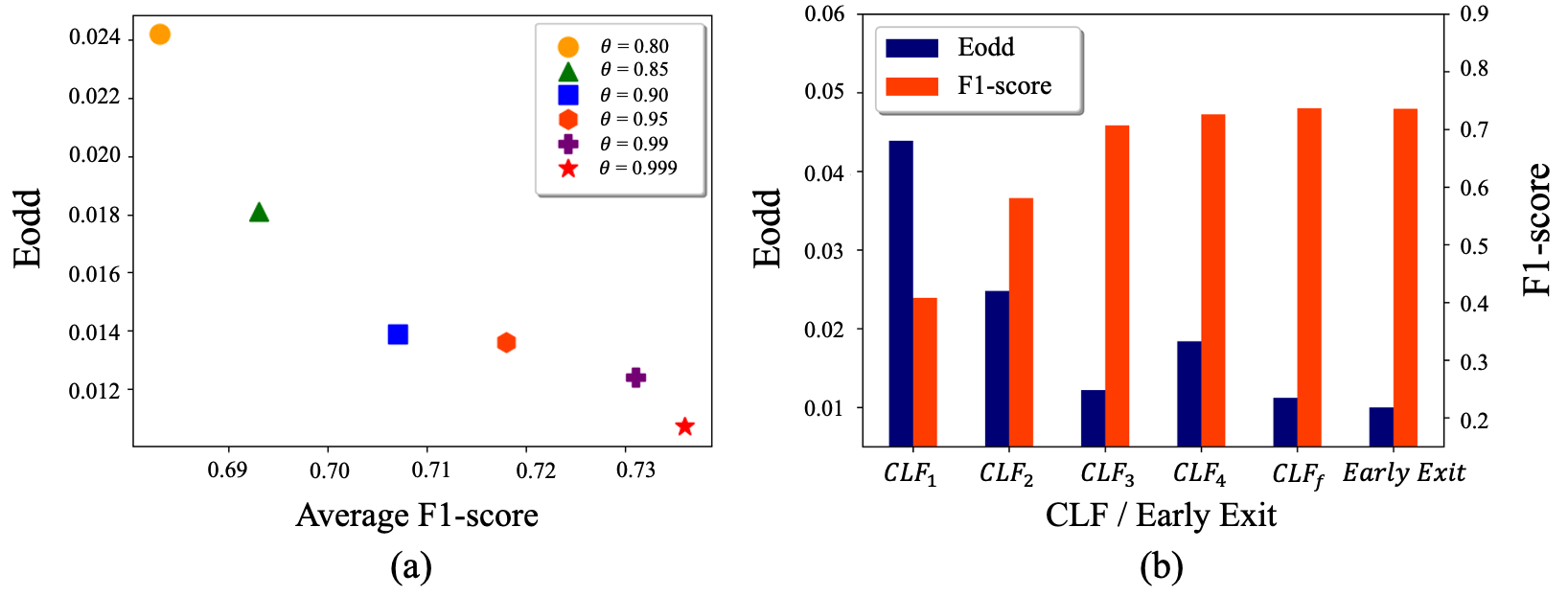}
\caption{Ablation study on (a) the early exit confidence threshold $\theta$ and (b) the early exit strategy for gender fairness on ISIC 2019 dataset. The $CLF_{1}$ through $CLF_{4}$ refer to internal classifiers, while $CLF_{f}$ refers to the final classifier.}
\label{Fig.Ablation}
\vspace{-20pt}
\end{center}
\end{figure}

\textbf{\indent Effect of Different Confidence Thresholds.} To investigate the impact of varying confidence thresholds $\theta$ on accuracy and fairness, we apply the ME-FairPrune method to a pre-trained model from the ISIC 2019 dataset and test different thresholds. Our results, shown in Fig. \ref{Fig.Ablation} (a), indicate that increasing the threshold improves accuracy and fairness. Thus, we recommend setting the threshold to 0.999 for optimal performance.

\textbf{Effect of Early Exits.} In Fig. \ref{Fig.Ablation} (b), we compare ME-FairPrune using an early exit policy with exiting from each specific exit. The results show that across all criteria, no specific $CLF$ outperforms the early exit policy in terms of Eodd, while our early exit policy achieves an accuracy level comparable to the original classifier output $CLF_{f}$. These findings underscore the importance of the proposed early exit strategy for achieving optimal performance.

%% file: Table/compare_fitz.tex
\begin{table}[h]
\centering
\caption{Results of accuracy and fairness of different methods on Fitzpatrick-17k dataset, using skin tone as the sensitive attribute. The dark skin is the privileged group with higher accuracy by vanilla training. 
The results of \emph{VGG-11}, \emph{AdvConf}, \emph{AdvRev}, \emph{DomainIndep} and \emph{FairPrune} are the experimental results reported in \cite{wu2022fairprune}.
}
\setlength{\tabcolsep}{6pt}
\label{Table:compare_fitz}
\scriptsize
\begin{tabular}{c c c c c c c c}\toprule
\multicolumn{1}{c}{\text{}} & \multicolumn{1}{c }{\text{}} &\multicolumn{3}{c }{\text{Accuracy}} & \multicolumn{3}{c }{\text{Fairness}}  \\ 
\cmidrule(lr){3-5}
\cmidrule(ll){6-8}
\multicolumn{1}{ c }{\text{Method}} & \text{Skin Tone} &\text{Precision} & \text{Recall} & \text{F1-score} & \text{Eopp0 $\downarrow$} & \text{Eopp1 $\downarrow$} & \text{Eodd $\downarrow$} \\ 
\hline
\hline
\multirow{4}{*}{VGG-11} & \text{Dark} & 0.563 & 0.581 & 0.546 & \multirow{4}{*}{0.0013} & \multirow{4}{*}{0.361} & \multirow{4}{*}{0.182} \\ 
 & \text{Light} & 0.482 & 0.495 & 0.473 &  &  & \\ 
 & \text{Avg. $\uparrow$} & 0.523 & 0.538 & 0.510 &  &  & \\ 
 & \text{Diff. $\downarrow$} & 0.081 & 0.086 & 0.073 &  &  & \\ 
\hline
\multicolumn{1}{ c }{\multirow{4}{*}{AdvConf}} & \text{Dark} & 0.506 & 0.562 & 0.506 & \multirow{4}{*}{0.0011} & \multirow{4}{*}{0.339} & \multirow{4}{*}{0.169} \\ 
 & \text{Light} & 0.427 & 0.464 & 0.426 &  &  & \\ 
 & \text{Avg. $\uparrow$} & 0.467 & 0.513 & 0.466 &  &  & \\ 
 & \text{Diff. $\downarrow$} & 0.079 & 0.098 & 0.080 &  &  & \\ 
\hline
\multicolumn{1}{ c }{\multirow{4}{*}{AdvRev}} & \text{Dark} & 0.514 & 0.545 & 0.503 & \multirow{4}{*}{0.0011} & \multirow{4}{*}{0.334} & \multirow{4}{*}{0.166} \\ 
 & \text{Light} & 0.489 & 0.469 & 0.457 &  &  & \\ 
 & \text{Avg. $\uparrow$} & 0.502 & 0.507 & 0.480 &  &  & \\ 
 & \text{Diff. $\downarrow$} & 0.025 & 0.076 & 0.046 &  &  & \\ 
\hline
\multicolumn{1}{ c }{\multirow{4}{*}{DomainIndep}} & \text{Dark} & 0.559 & 0.540 & 0.530 & \multirow{4}{*}{0.0012} & \multirow{4}{*}{0.323} & \multirow{4}{*}{0.161} \\ 
 & \text{Light} & 0.541 & 0.529 & 0.512 &  &  & \\ 
 & \text{Avg. $\uparrow$} & 0.550 & 0.534 & 0.521 &  &  & \\ 
 & \text{Diff. $\downarrow$} & 0.018 & 0.010 & 0.018 &  &  & \\ 
\hline
\multicolumn{1}{ c }{\multirow{4}{*}{HSIC}} & \text{Dark} & 0.548 & 0.522 & 0.513 & \multirow{4}{*}{0.0013} & \multirow{4}{*}{0.331} & \multirow{4}{*}{0.166} \\ 
 & \text{Light} & 0.513 & 0.506 & 0.486 &  &  & \\ 
 & \text{Avg.$\uparrow$} & 0.530 & 0.515 & 0.500 &  &  & \\ 
 & \text{Diff. $\downarrow$} & 0.040 & 0.018 & 0.029 &  &  & \\ 
\hline

\multicolumn{1}{ c }{\multirow{4}{*}{MFD}} & \text{Dark} & 0.514 & 0.545 & 0.503 & \multirow{4}{*}{0.0011} & \multirow{4}{*}{0.334} & \multirow{4}{*}{0.166} \\ 
 & \text{Light} & 0.489 & 0.469 & 0.457 &  &  & \\ 
 & \text{Avg. $\uparrow$} & 0.502 & 0.507 & 0.480 &  &  & \\ 
 & \text{Diff. $\downarrow$} & 0.025 & 0.076 & 0.046 &  &  & \\ 
\hline
\hline
{\multirow{4}{*}{FairPrune}} & \text{Dark} & 0.567 & 0.519 & 0.507 & \multirow{4}{*}{\textbf{0.0008}} & \multirow{4}{*}{0.330} & \multirow{4}{*}{0.165} \\ 
 & \text{Light} & 0.496 & 0.477 & 0.459 &  &  & \\ 
 & \text{Avg. $\uparrow$} & 0.531 & 0.498 & 0.483 &  &  & \\ 
 & \text{Diff. $\downarrow$} & 0.071 & 0.042 & 0.048 &  &  & \\ 
\hline
\multicolumn{1}{ c }{\multirow{4}{*}{ME-FairPrune}} & \text{Dark} & 0.564 & 0.529 & 0.523 & \multirow{4}{*}{0.0012} & \multirow{4}{*}{\textbf{0.305}} & \multirow{4}{*}{\textbf{0.152}} \\ 
 & \text{Light} & 0.542 & 0.535 & 0.522 &  &  & \\ 
 & \text{Avg. $\uparrow$} & 0.553 & 0.532 & 0.522 &  &  & \\ 
 & \text{Diff. $\downarrow$} & 0.022 & 0.006 & 0.001 &  &  & \\
\bottomrule
\end{tabular}
\end{table}


%% file: Table/compare_isic_me.tex
\begin{table}[h]
\centering
\caption{Accuracy and fairness of classification results across different baselines with and without the ME training framework on the ISIC 2019 dataset.}
\setlength{\tabcolsep}{6pt}
\label{Table:compare_isic_me}
\scriptsize
\begin{tabular}{c c c c c c c c}\toprule
\multicolumn{1}{c }{\text{}} & \multicolumn{1}{c }{\text{}} &\multicolumn{3}{c }{\text{Accuracy}} & \multicolumn{3}{c }{\text{Fairness}}  \\ 
\cmidrule(lr){3-5}
\cmidrule(ll){6-8}
\multicolumn{1}{ c }{\text{Method}} & \text{Gender} &\text{Precision} & \text{Recall} & \text{F1-score} & \text{Eopp0 $\downarrow$} & \text{Eopp1 $\downarrow$} & \text{Eodd $\downarrow$} \\ 
\hline
\hline

\multicolumn{1}{ c }{\multirow{4}{*}{ResNet18}} & \text{Female} & 0.793 & 0.721 & 0.746 & \multirow{4}{*}{\textbf{0.006}} & \multirow{4}{*}{0.044} & \multirow{4}{*}{0.022} \\ 
 & \text{Male} & 0.731 & 0.725 & 0.723 &  &  & \\ 
 & \text{Avg. $\uparrow$} & 0.762 & 0.723 & 0.735 &  &  & \\ 
 & \text{Diff. $\downarrow$} & 0.063 & 0.004 & 0.023 &  &  & \\ 
\hline

\multicolumn{1}{ c }{\multirow{4}{*}{ME-ResNet18}} & \text{Female} & 0.748 & 0.724 & 0.733 & \multirow{4}{*}{\textbf{0.006}} & \multirow{4}{*}{\textbf{0.031}} & \multirow{4}{*}{\textbf{0.016}} \\ 
 & \text{Male} & 0.723 & 0.736 & 0.726 &  &  & \\ 
 & \text{Avg. $\uparrow$} & 0.735 & 0.730 & 0.730 &  &  & \\ 
 & \text{Diff. $\downarrow$} & 0.025 & 0.012 & 0.007 &  &  & \\ 
\hline
\hline

\multicolumn{1}{ c }{\multirow{4}{*}{HSIC}} & \text{Female} & 0.744 & 0.660 & 0.696 & \multirow{4}{*}{0.008} & \multirow{4}{*}{0.042} & \multirow{4}{*}{0.020} \\ 
 & \text{Male} & 0.718 & 0.697 & 0.705 &  &  & \\ 
 & \text{Avg. $\uparrow$} & 0.731 & 0.679 & 0.700 &  &  & \\ 
 & \text{Diff. $\downarrow$} & 0.026 & 0.037 & 0.009 &  &  & \\ 
\hline

\multicolumn{1}{ c }{\multirow{4}{*}{ME-HSIC}} & \text{Female} & 0.733 & 0.707 & 0.716 & \multirow{4}{*}{\textbf{0.007}} & \multirow{4}{*}{{\textbf{0.034}}} & \multirow{4}{*}{{\textbf{0.018}}} \\ 
 & \text{Male} & 0.713 & 0.707 & 0.707 &  &  & \\ 
 & \text{Avg. $\uparrow$} & 0.723 & 0.707 & 0.712 &  &  & \\ 
 & \text{Diff. $\downarrow$} & 0.020 & 0.000 & 0.009 &  &  & \\ 
\hline
\hline

\multicolumn{1}{ c }{\multirow{4}{*}{MFD}} & \text{Female} & 0.770 & 0.697 & 0.726 & \multirow{4}{*}{\textbf{0.005}} & \multirow{4}{*}{0.051} & \multirow{4}{*}{0.024} \\ 
 & \text{Male} & 0.772 & 0.726 & 0.744 &  &  & \\ 
 & \text{Avg. $\uparrow$} & 0.771 & 0.712 & 0.735 &  &  & \\ 
 & \text{Diff. $\downarrow$} & 0.002 & 0.029 & 0.018 &  &  & \\ 
\hline

\multicolumn{1}{ c }{\multirow{4}{*}{ME-MFD}} & \text{Female} & 0.733 & 0.698 & 0.711 & \multirow{4}{*}{\textbf{0.005}} & \multirow{4}{*}{\textbf{0.024}} & \multirow{4}{*}{\textbf{0.012}} \\ 
 & \text{Male} & 0.772 & 0.713 & 0.739 &  &  & \\ 
 & \text{Avg. $\uparrow$} & 0.752 & 0.706 & 0.725 &  &  & \\ 
 & \text{Diff. $\downarrow$} & 0.039 & 0.015 & 0.028 &  &  & \\ 

\bottomrule
\end{tabular}
\end{table}


%% file: Section/Conclusion.tex
We address the issue of deteriorating fairness in deeper layers of deep neural networks by proposing a multi-exit training framework. Our framework can be applied to various bias mitigation methods and uses a confidence-based exit strategy to simultaneously achieve high accuracy and fairness. Our results demonstrate that our framework achieves the best trade-off between accuracy and fairness compared to the state-of-the-art on two dermatological disease datasets.


%% file: Table/supp_compare_fitz_me.tex
\begin{table*}[t]
\centering
\caption{Accuracy and fairness of classification results across different baselines with and without the ME training framework on the Fitzpatrick-17 dataset. The backbone for these experiments are VGG-11.}
\setlength{\tabcolsep}{6pt}
\label{Table:compare_fitz_me}
\scriptsize
\begin{tabular}{c c c c c c c c}\toprule
\multicolumn{1}{c }{\text{}} & \multicolumn{1}{c }{\text{}} &\multicolumn{3}{c }{\text{Accuracy}} & \multicolumn{3}{c }{\text{Fairness}}  \\ 
\cmidrule(lr){3-5}
\cmidrule(ll){6-8}
\multicolumn{1}{ c }{\text{Method}} & \text{Skin Tone} &\text{Precision} & \text{Recall} & \text{F1-score} & \text{Eopp0 $\downarrow$} & \text{Eopp1 $\downarrow$} & \text{Eodd $\downarrow$} \\ 
\hline
\hline

\multirow{4}{*}{VGG-11} & \text{Dark} & 0.563 & 0.581 & 0.546 & \multirow{4}{*}{0.0013} & \multirow{4}{*}{0.361} & \multirow{4}{*}{0.182} \\ 
 & \text{Light} & 0.482 & 0.495 & 0.473 &  &  & \\ 
 & \text{Avg. $\uparrow$} & 0.523 & 0.538 & 0.510 &  &  & \\ 
 & \text{Diff. $\downarrow$} & 0.081 & 0.086 & 0.073 &  &  & \\ 
\hline

\multirow{4}{*}{ME-VGG-11} & \text{Dark} & 0.568 & 0.517 & 0.517 & \multirow{4}{*}{\textbf{0.0012}} & \multirow{4}{*}{\textbf{0.320}} & \multirow{4}{*}{\textbf{0.160}} \\ 
 & \text{Light} & 0.522 & 0.536 & 0.514 &  &  & \\ 
 & \text{Avg. $\uparrow$} & 0.545 & 0.527 & 0.516 &  &  & \\ 
 & \text{Diff. $\downarrow$} & 0.046 & 0.019 & 0.003 &  &  & \\ 
\hline
\hline

\multicolumn{1}{ c }{\multirow{4}{*}{HSIC}} & \text{Dark} & 0.548 & 0.522 & 0.513 & \multirow{4}{*}{0.0013} & \multirow{4}{*}{0.331} & \multirow{4}{*}{0.166} \\ 
 & \text{Light} & 0.513 & 0.506 & 0.486 &  &  & \\ 
 & \text{Avg.$\uparrow$} & 0.530 & 0.515 & 0.500 &  &  & \\ 
 & \text{Diff. $\downarrow$} & 0.040 & 0.018 & 0.029 &  &  & \\ 
\hline

\multicolumn{1}{ c }{\multirow{4}{*}{ME-HSIC}} & \text{Dark} & 0.566 & 0.528 & 0.528 & \multirow{4}{*}{\textbf{0.0012}} & \multirow{4}{*}{{\textbf{0.314}}} & \multirow{4}{*}{{\textbf{0.157}}} \\ 
 & \text{Light} & 0.523 & 0.518 & 0.503 &  &  & \\ 
 & \text{Avg. $\uparrow$} & 0.544 & 0.523 & 0.515 &  &  & \\ 
 & \text{Diff. $\downarrow$} & 0.043 & 0.010 & 0.025 &  &  & \\ 
\hline
\hline

\multicolumn{1}{ c }{\multirow{4}{*}{MFD}} & \text{Dark} & 0.514 & 0.545 & 0.503 & \multirow{4}{*}{\textbf{0.0011}} & \multirow{4}{*}{0.334} & \multirow{4}{*}{0.166} \\ 
 & \text{Light} & 0.489 & 0.469 & 0.457 &  &  & \\ 
 & \text{Avg. $\uparrow$} & 0.502 & 0.507 & 0.480 &  &  & \\ 
 & \text{Diff. $\downarrow$} & 0.025 & 0.076 & 0.046 &  &  & \\ 
\hline

\multicolumn{1}{ c }{\multirow{4}{*}{ME-MFD}} & \text{Dark} & 0.561 & 0.545 & 0.535 & \multirow{4}{*}{0.0012} & \multirow{4}{*}{\textbf{0.319}} & \multirow{4}{*}{\textbf{0.159}} \\ 
 & \text{Light} & 0.539 & 0.528 & 0.512 &  &  & \\ 
 & \text{Avg. $\uparrow$} & 0.550 & 0.537 & 0.523 &  &  & \\ 
 & \text{Diff. $\downarrow$} & 0.022 & 0.017 & 0.023 &  &  & \\ 

\bottomrule
\end{tabular}
\end{table*}